# Towards Automated Diagnosis of Inherited Arrhythmias: Combined Arrhythmia Classification Using Lead-Aware Spatial Attention Networks

**Authors:** Sophie Sigfstead[1], River Jiang[2] MD, Brianna Davies[2] MSc, Zachary W.M. Laksman[2] MD, Julia Cadrin-Tourigny[3] MD PhD, Rafik Tadros[3] MD PhD, Habib Khan[5] MD PhD, Joseph Atallah[1] MD MSc, Christian Steinberg[6] MD, Shubhayan Sanatani[7] MD, Mario Talajic[4] MD, Rahul Krishnan[7] PhD, Andrew D. Krahn[2] MD, Christopher C. Cheung[7,8] MD MPH*

**Affiliations:**

1. Department of Mathematical and Statistical Sciences, University of Alberta, Edmonton, AB, Canada.

2. Division of Cardiology, University of British Columbia, Vancouver, BC, Canada.

3. Montreal Heart Institute, Montreal, QC, Canada.

4. London Health Sciences Center, London, ON, Canada.

5. Quebec Heart and Lung Institute, Quebec, QC, Canada.

6. Children's Heart Centre, Vancouver, BC, Canada.

7. University of Toronto, Toronto, ON, Canada

8. Division of Cardiology, Sunnybrook Health Sciences Centre, Toronto, ON, Canada.

*Correspondence: christopher.cheung@sunnybrook.ca

## 0. ABSTRACT

**Background:** Arrhythmogenic right ventricular cardiomyopathy (ARVC) and long QT syndrome (LQTS) are inherited arrhythmia syndromes associated with sudden cardiac death. Deep learning shows promise for ECG interpretation, but multi-class inherited arrhythmia classification with clinically grounded interpretability remains underdeveloped.

**Objective**: To develop and validate a lead-aware deep learning framework for multi-class (ARVC vs LQTS vs control) and clinically relevant binary inherited arrhythmia classification, and to determine optimal strategies for integrating ECG foundation models within arrhythmia screening tools.

**Methods:** We assembled a 13-center Canadian cohort (645 patients; 1,344 ECGs). We evaluated four ECG foundation models using three transfer learning approaches: linear probing, fine-tuning, and combined strategies. We developed lead-aware spatial attention networks (LASAN) and assessed integration strategies combining LASAN with foundation models. Performance was compared against the established foundation model baselines. Lead-group masking quantified disease-specific lead dependence.

**Results:** Fine-tuning outperformed linear probing across all foundation models (mean macro-AUROC 0.904 vs 0.825). The best lead-aware integrations achieved near-ceiling performance (HuBERT-ECG hybrid: macro-AUROC 0.990; ARVC vs control AUROC 0.999; LQTS vs control AUROC 0.994). Lead masking demonstrated physiologic plausibility: V1–V3 were most critical for ARVC detection (4.54% AUROC reduction), while lateral leads were preferentially important for LQTS (2.60% drop).

**Conclusion:** Lead-aware architectures achieved state-of-the-art performance for inherited arrhythmia classification, outperforming all existing published models on both binary and multi-class tasks while demonstrating clinically aligned lead dependence. These findings support potential utility for automated ECG screening and triage pending prospective validation.

1. **INTRODUCTION**

Arrhythmogenic right ventricular cardiomyopathy (ARVC) and long QT syndrome (LQTS) represent inherited arrhythmia syndromes that disproportionately affect young adults, carrying substantial risk of sudden cardiac death.[1, 2] ARVC, characterized by progressive fibro-fatty replacement of right ventricular myocardium, affects approximately 1 in 2,000 to 1 in 5,000 individuals and accounts for up to 20% of sudden cardiac death cases in individuals under 35 years.[3-6] LQTS, arising from ion channel dysfunction that prolongs cardiac repolarization, has a prevalence of approximately 1 in 2,000 and predisposes affected individuals to torsades de pointes and ventricular fibrillation.[7, 8] Diagnosis of both conditions relies on multi-parameter criteria incorporating electrocardiographic findings, cardiac imaging, and genetic testing.[1, 3, 8] Early arrhythmia diagnosis enables life-saving interventions including implantable cardioverter-defibrillators, beta-blocker therapy, lifestyle modifications, and family screening.[8]

The 12-lead electrocardiogram (ECG) serves as the first-line screening tool for arrhythmia given its accessibility and cost-effectiveness, yet diagnostic sensitivity remains suboptimal in ARVC and LQTS. The 2010 Task Force Criteria for ARVC incorporate ECG features are limited to achieve 88% sensitivity and 73% specificity using standard ECG.[9] Similarly, LQTS screening via QTc measurement demonstrates sensitivity of approximately 72%, with significant inter-reader variability.[10] These limitations underscore the need for automated, consistent, high-sensitivity screening approaches.

Deep learning has achieved remarkable accuracy for ECG classification in various electrophysiology contexts, with cardiologist-level performance demonstrated across multiple condition types.[11-14] However, within inherited arrhythmia, most prior work in has focused on single-disease detection (binary classification).[11, 15-17] Robust, clinically deployable models capable of differentiating multiple inherited syndromes within a unified framework remain underdeveloped, representing a critical gap, as clinical screening scenarios often require distinguishing among diverse arrhythmia presentations.[11, 15, 16, 18] Additionally, deployment of such models requires not only high discriminative performance but also clinically grounded interpretability to ensure model reasoning aligns with known electrophysiological signatures.[19, 20]

Foundation models pretrained on large ECG datasets have demonstrated promise for transfer learning across diverse cardiovascular conditions.[14, 21-24] Yet the optimal transfer learning strategy for integrating these representations in inherited arrhythmia detection remains unclear, with questions persisting about whether linear probing, end-to-end fine-tuning, or hybrid approaches yield superior performance.[21, 22, 25-27] Separately, the interpretability of models is important, and models that rely on physiologically plausible signals can improve clinicians trust.[20, 28] Given that each arrhythmia presents distinct electrophysiological signatures in specific leads (e.g., V1–V3 reflecting right ventricular activity in ARVC; lateral leads capturing repolarization features in LQTS), lead-level importance patterns can serve as a clinically meaningful validation of model reasoning.[11, 29-34]

Accordingly, this study addresses three core questions. First, can a deep learning architecture achieve state-of-the-art performance for both binary and multi-class inherited arrhythmia classification across real-world, multi-center data? Second, how should ECG foundation models be integrated for this task—do existed pre-trained representations, end-to-end adaption, or a combined approach (linear probing and fine tuning) provide the most meaningful gains, and how can these architectures be incorporated into model pipelines? Third, do learned feature representations demonstrate clinically interpretable lead-dependence patterns consistent with known disease electrophysiology?

To answer these questions, we developed lead-aware spatial attention networks (LASAN), an architecture designed to explicitly model per-lead cardiac anatomy and capture disease-specific spatial patterns across the 12-lead ECG. [30, 31, 35] We systematically evaluated four ECG foundation models under linear probing, fine-tuning, and combination (linear probing followed by fine tuning) strategies, assessing both standalone LASAN performance and optimal approaches for incorporating pretrained representations. Using ECGs from thirteen Canadian electrophysiology centers, we benchmarked performance on multi-class classification (ARVC vs LQTS vs control) and clinically relevant binary detection tasks (ARVC vs control; LQTS vs control). Finally, we employed lead-group masking to quantify disease-specific lead dependence, providing interpretable validation that model predictions align with established electrophysiological principles.

## 2. METHODS

### 2.1 Data Collection

ECG records were sourced from the Canadian Hearts in Rhythm Organization (HiRO) ARVC Registry, which aggregates data from 13 Canadian cardiac centers and includes individuals with confirmed arrhythmogenic right ventricular cardiomyopathy (ARVC), long QT syndrome (LQTS), and controls.

ARVC diagnosis was established according to current Task Force Criteria, including patients diagnosed with definite or probable ARVC based on clinical, electrocardiographic, imaging, and genetic criteria.[1] Patients with possible ARVC diagnosis were excluded. ARVC Patients were categorized as gene-positive (PKP2, DSG2, DSC2, DSP, JUP, or TMEM43 pathogenic/likely pathogenic variants) or gene-negative based on genetic testing.[1] LQTS diagnosis required identification of a pathogenic/likely pathogenic variant in LQTS-associated genes (KCNQ1, KCNH2, or SCN5A) with categorization as Long QT Syndrome Type 1 (LQT1) or Type 2 (LQT2) based on genetic subtype, or clinician diagnosis.[8] Control participants had no known inherited cardiac arrhythmia but were not limited to normal electrocardiograms to mimic clinical conditions. Standard 12-lead electrocardiograms were acquired during routine clinical care using institutional electrocardiography systems, with sampling rates of 250 Hz and 10-second duration. ECGs containing only non-augmented leads (8-lead) were stored in XML file format, which were used for training and inference in downstream ML models.

## 2.2 Data Preprocessing

Eight-lead electrocardiographic signals were extracted from institutional databases and processed using a standardized data processing pipeline. Per-lead z-score normalization was applied to account for inter-individual amplitude variations and equipment calibration differences, with each lead normalized to zero mean and unit variance across the 2,500-sample window. An electrophysiologist (C.C.) reviewed each ECG to verify signal integrity, and ECGs with significant artifact, incomplete recordings, or technical errors were excluded. Preprocessed signals were stored as 32-bit floating-point tensors (8 leads x 2,500 samples). To match foundation model input requirements, ECGs were also up-sampled to 500 Hz, truncated to a 5-second duration and stored in the same format (8 leads x 2500 samples).

**2.3 Data Splitting**

Patient-level, stratified splitting was used to prevent data leakage and to provide robust performance estimates. All ECG recordings from a given patient were assigned to a single partition (training, validation, or test) across all analyses. Patients were allocated in a 70:15:15 ratio to the training, validation, and test sets using a fixed random seed (seed = 42) to ensure reproducibility. Site-based splitting was not performed to avoid undersized partitions for certain tasks and to prevent lead-wise encoding approaches from being confounded by site-specific recording signatures. In addition, some patients underwent ECG acquisition at multiple sites, making it challenging to guarantee strict site separation without reintroducing patient overlap; therefore, patient-level separation was prioritized. Within each diagnostic group (ARVC, LQTS, Control), stratified sampling preserved proportional class representation across partitions, with no patient contributing ECGs to more than one set. As our main objective is to inform screening tools, this strategy evaluates generalization to previously unseen patients and provides an assessment of clinical utility for the intended use case.

**2.4 Model Development**

**2.4.1 Foundation Model Baselines**

We evaluated four state-of-the-art electrocardiogram foundation models as baseline comparisons, each pretrained on large-scale ECG databases and with varied underlying architectures:

1. **ECG-Founder**: A ResNet-based architecture pretrained on 10.8 million ECGs with 150 diagnostic categories from the Harvard-Emory ECG Database.[21]
2. **ECG-FM**: A transformer-based foundation model pretrained on 1.5 million ECGs using hybrid contrastive and generative self-supervised learning.[22]
3. **HuBERT-ECG**: A self-supervised transformer encoder pretrained on 9.1 million 12-lead ECGs encompassing 164 cardiovascular conditions using masked prediction of hidden units.[24]
4. **Deep-ECG-SSL**: An EfficientNet-based model employing self-supervised contrastive learning and masked lead modeling, trained on over 1 million ECGs and validated on 77 cardiac classification categories.[23]

For each foundation model, we compared three training strategies to determine optimal transfer learning approaches and provide optimal performance baselines:[36, 37]

1. **Linear Probing**: The pretrained encoder was frozen, and only a linear classification head was trained for 50 epochs using a learning rate of $1\times10^{-2}$ with cosine annealing schedule and weight decay of $1\times10^{-4}$.
2. **Fine-Tuning**: All model parameters were updated for 100 epochs using a learning rate of $1\times10^{-4}$, weight decay of $1\times10^{-4}$, and gradient clipping (maximum norm = 1.0) to prevent catastrophic forgetting of pretrained representations.
3. **Linear Probing then Fine-Tuning**: A combined approach that first trains a linear classifier on frozen features (50 epochs) before fine-tuning the entire model (100 additional epochs), a method which has shown to preserve pretrained features while enabling adaptation to downstream tasks.[36]

**2.4.2 Lead-Aware Spatial Attention Networks (LASAN) Architectures**

Within the electrocardiogram, specific leads are critical for different arrhythmias. For example, leads V1-V3 are directly related to right ventricular activity (essential for ARVC detection), while lateral leads (e.g., I, V5-V6) capture left ventricular repolarization patterns (critical for LQTS detection).[32-34]

To explicitly model the anatomical organization of electrocardiographic leads, we developed a novel architecture incorporating lead-specific feature extraction, positional encoding, and inter-lead attention mechanisms. The LASAN architecture consists of five primary components (**Figure 1a**):

1. Per-Lead Temporal Encoder: Each of the 8 input leads were processed independently through a 1-dimensional convolutional neural network to extract lead-specific temporal features. The encoder comprised four convolutional blocks with progressively increasing channel dimensions (32, 64, 128, 256), kernel size of 15 samples, batch normalization, ReLU activation, and max pooling (stride=2), producing a 256-dimensional feature vector per lead.
2. Anatomical Position Encoding: Learnable position embeddings encoding both lead identity and lead grouping were incorporated.[30] Each lead was assigned a 256-dimensional trainable embedding vector added to the per-lead features, enabling the model to learn anatomical distinctions between specific lead groups.[30]
3. Inter-Lead Transformer: As standard three-layer transformer encoder with 4 attention heads, 256 hidden dimension, 512-dimensional feed forward layers, and 0.1 dropout modeled dependencies between leads.[38] Multi-head self-attention mechanisms computed attention scores between all pairs of leads, allowing the model to learn clinically meaningful lead relationships (e.g., V1-V3 co-activation in ARVC, concordant lateral lead changes in LQTS).

4. Lead Importance Aggregator: A single-head attention mechanism with learned query vector produced interpretable per-lead importance weights (8-dimensional vector) and a weighted 256-dimensional aggregate representation. These attention weights quantify each lead's contribution to the final classification decision, enabling post-hoc interpretability through lead masking experiments.

5. Classification Head: A two-layer multi-layer perceptron with hidden dimensions (256 to 128), ReLU activation, 0.25 dropout, and softmax output produced class probabilities for 3-class classification (ARVC, LQTS, Control). In binary classification tasks, a sigmoid function was used to provide outputs (**Section 3.4**).

**2.4.3 Model Integration Strategies**

We evaluated three approaches for integrating LASAN networks with foundation models:

1. **Standalone LASAN (Figure 1b)**: A complete LASAN architecture trained from random initialization.

2. **Foundation LASAN Head (Figure 1c)**: Fine-tuned foundation model encoder with LASAN's lead importance aggregator and classification head replacing a typical standard linear probe.

3. **Hybrid LASAN (Figure 1d)**: A dual-branch architecture combining the outputs of a foundation model features and trainable LASAN architecture, fused together using a gated mechanism before the classification head.[39]

### 2.4.4 Training Configuration

All models were optimized using focal loss to address class imbalance, focusing the model on learning more difficult to classify examples[40]. Training employed the Adam optimizer with batch size 32, cosine annealing learning rate schedule, 5-epoch warmup, and early stopping based on validation AUROC (patience: 15 epochs for 3-class classification, 20 epochs for binary tasks). Gradient clipping (maximum norm = 1.0) reduced training instability. Hyperparameters were selected through systematic grid search on the validation set: learning rates ($1 \times 10^{-5}$ to $1 \times 10^{-2}$ for linear probing; $1 \times 10^{-5}$ to $1 \times 10^{-3}$ for fine-tuning), weight decay ($1 \times 10^{-5}$ to $1 \times 10^{-3}$), and dropout rates (0.1 to 0.3). Models were trained using parallelization on 8 NVIDIA A100 GPUs.

### 2.5 Evaluation Metrics

The primary outcome was area under the receiver operating characteristic curve (AUROC) for both 3-class (macro-averaged) and binary classification tasks. Secondary metrics included sensitivity (recall), specificity, and balanced accuracy. All analyses were performed using Python 3.10 with scikit-learn 1.3 and SciPy 1.11.

## 3. RESULTS

### 3.1 Patient Demographics

The study cohort comprised 645 patients across 13 Canadian cardiac centers, including 121 ARVC patients, 268 LQTS patients, and 256 controls (**Table 1**). The dataset contained 1,344 electrocardiograms, with a mean of 3.13 ECGs per ARVC patient, 1.74 per LQTS patient, and 1.95 per control.

Among ARVC patients, 74 (61.2%) carried pathogenic variants with 53 meeting definite and 21 meeting probable diagnostic criteria. Gene-negative ARVC patients (n=47) were classified as definite (n=27) or probable (n=20) based on clinical criteria. Mean ARVC heart rate was $65.7 \pm 14.8$ bpm, with QTc of $449.1 \pm 50.6$ ms, PR interval of $167.7 \pm 42.8$ ms, and QRS duration of $113.1 \pm 32.5$ ms. LQTS patients included 194 with LQT1 (KCNQ1 variants) and 74 with LQT2 (KCNH2 variants), with median age of 34.9 years (range 0.26–85.77). Mean QT interval was $435.32 \pm 53.10$ ms across the LQTS cohort. Control participants had a mean age of $38.5 \pm 17.0$ years (range 2.0–80.0). Mean control QT interval was $396.3 \pm 36.7$ ms with heart rate of $69.7 \pm 14.1$ bpm.

### 3.2 Foundation Model Training Strategy Performance

We first evaluated four ECG foundation models (ECG-Founder, Deep-ECG-SSL, ECG-FM, HuBERT-ECG) under three transfer learning strategies (linear probing, fine tuning and linear-probing then fine tuning, as described previously) to establish baseline performance on the 3-class task (ARVC vs LQTS vs Control) (**Table 2**).

Linear probing yielded a mean macro-AUROC of 0.825 across models (range 0.807–0.852). Performance varied by foundation model, with ECG-Founder achieving the highest AUROC under linear probing (0.852) and HuBERT-ECG the lowest (0.807). Consistent with this variability, the reported AUROC standard deviations under linear probing were comparatively larger (0.035–0.056 across models), indicating less stable optimization when relying solely on frozen foundation model representations.

In contrast, end-to-end fine-tuning improved performance across all four foundation models, producing a mean macro-AUROC of 0.904 (range 0.889–0.910). The top-performing fine-tuned models were ECG-Founder (0.910), ECG-FM (0.909), and Deep-ECG-SSL (0.908), while HuBERT-ECG also improved to 0.889. Fine-tuning also reduced dispersion in performance across models and yielded smaller AUROC standard deviations (0.013–0.025), consistent with more stable convergence when pretrained representations were permitted to adapt to the inherited arrhythmia task. The two-stage "linear probe then fine-tune" approach achieved intermediate-to-comparable performance (mean macro-AUROC 0.898, range 0.883–0.909) and similarly low variability (standard deviations 0.014–0.025).

Per-class analyses demonstrated that the benefit of fine-tuning was present across all conditions, with the largest gains observed for ARVC. Averaged across foundation models, mean ARVC AUROC increased from 0.828 under linear probing to 0.925 with fine-tuning (+0.098), whereas LQTS improved from 0.845 to 0.915 (+0.070) and Control improved from 0.803 to 0.873 (+0.070) (Table 2). Under fine-tuning, all models achieved strong class-wise discrimination (ARVC AUROC 0.907–0.942; LQTS AUROC 0.898–0.927; Control AUROC 0.862–0.882).

Collectively, these findings indicate that task-specific adaptation is important for inherited arrhythmia classification and that frozen feature representations incompletely capture the disease-specific morphology required for robust multi-class discrimination in this setting.

**3.3 LASAN Model Performance**

We next evaluated LASAN architectures to test whether explicit lead-aware modeling improves performance beyond foundation-model baselines (**Table 3; Figure 1**). Standalone LASAN trained from random initialization achieved a macro-AUROC of 0.911 ± 0.037 on the 3-class task, with balanced class-wise performance (LQTS 0.932; ARVC 0.940; Control 0.862). Notably, standalone LASAN exceeded the best linear-probed foundation model (0.852) and approached the performance of fine-tuned foundation models (0.889–0.910), supporting the utility of lead-aware inductive bias for capturing spatially localized electrophysiologic signatures.

When integrating LASAN by replacing the standard classifier with a LASAN head on top of frozen foundation encoders ("foundation model with LASAN head"), performance was generally strong but varied across foundation models (**Table 3**). Macro-AUROC ranged from 0.827 ± 0.097 (ECG-Founder) to 0.990 ± 0.003 (HuBERT-ECG). Deep-ECG-SSL with a LASAN head achieved 0.909 ± 0.014, comparable to standalone LASAN (0.911) and to its fine-tuned baseline (0.908). Notably, ECG-FM and HuBERT-ECG paired particularly well with the LASAN head, achieving macro-AUROC of 0.973 ± 0.004 and 0.990 ± 0.003, respectively, with consistently high per-class discrimination (ECG-FM: LQTS 0.983, ARVC 0.971, Control 0.964; HuBERT-ECG: LQTS 0.989, ARVC 0.994, Control 0.987). ECG-Founder with a LASAN head achieved 0.827 ± 0.097, indicating that performance in this configuration can be sensitive to the pretrained representation used. These results also demonstrate that a combination of lead aware processing and foundation model utilization can deliver excellent multi-class performance.

Given the above findings, we next evaluated whether branched architecture, including both LASAN and foundational models (rather than stacked configuration) could also provide high performance, leading to development of the Hybrid LASAN architecture. This architecture fuses trainable lead-specific LASAN encoders with foundation model features, delivered consistently strong performance across all evaluated foundation models (**Table 3**). Hybrid LASAN achieved macro-AUROC values of 0.950 ± 0.011 (ECG-Founder), 0.932 ± 0.016 (Deep-ECG-SSL), 0.975 ± 0.003 (ECG-FM), and 0.990 ± 0.002 (HuBERT-ECG). Relative to the "foundation model with LASAN head" configuration, the hybrid formulation provided more uniform performance across pretrained encoders while maintaining the highest overall results for HuBERT-ECG (0.990) and ECG-FM (0.975). Class-wise AUROCs remained high and balanced (e.g., HuBERT-ECG hybrid: LQTS 0.989, ARVC 0.994, Control 0.990), and Control discrimination improved compared with standalone LASAN (0.862) across multiple hybrid variants. Overall, these findings suggest that combining global foundation representations with lead-aware, task-trained processing can yield both high performance and greater robustness to the choice of pretrained encoder.

### 3.4 Binary Classification

To assess clinically aligned use cases and provide comparison to existing detection models, we evaluated binary tasks reflecting screening and differential diagnosis scenarios (**Table 4**). For ARVC vs Control, standalone LASAN achieved AUROC 0.974 with perfect sensitivity (1.000), but lower specificity (0.864), yielding accuracy 0.941. Hybrid models improved overall discrimination and/or specificity, with HuBERT-ECG hybrid achieving AUROC 0.999, sensitivity 0.961, specificity 0.983, and accuracy 0.971. ECG-FM hybrid achieved AUROC 0.991 with specificity 0.966, and ECG-Founder hybrid achieved AUROC 0.978 with specificity 0.924. These results indicate excellent ARVC screening performance across architectures, with the highest-performing hybrid model achieving near-ceiling discrimination while maintaining high specificity.

For LQTS vs Control, standalone LASAN demonstrated more modest discrimination (AUROC 0.901), driven by limited sensitivity (0.701) despite high specificity (0.939), consistent with the known subtlety and phenotypic heterogeneity of repolarization abnormalities. Hybrid integration improved performance, particularly for HuBERT-ECG (AUROC 0.994, sensitivity 0.948, specificity 0.970, accuracy 0.958) and ECG-FM (AUROC 0.962, specificity 0.955). ECG-Founder hybrid achieved AUROC 0.939 with sensitivity 0.870 but lower specificity (0.848), whereas Deep-ECG-SSL hybrid achieved AUROC 0.930 with reduced specificity (0.720). Collectively, these results show that foundation-feature integration is particularly beneficial for LQTS screening, improving sensitivity while maintaining high specificity in the best-performing models.

Finally, for LQTS genotype discrimination (LQT1 vs LQT2), performance was lower than disease-vs-control tasks across models (**Table 4**). Standalone LASAN achieved AUROC 0.920 for genotype differentiation. Among hybrid models, Deep-ECG-SSL achieved the highest AUROC (0.948) with sensitivity 0.977 and specificity 0.650, while HuBERT-ECG achieved

AUROC 0.901 (sensitivity 0.841, specificity 0.775). ECG-Founder (0.764) and ECG-FM (0.738) performed comparatively poorly on this subtype task. Overall, binary task results preserved the hierarchy observed in multi-class experiments (HuBERT-ECG and ECG-FM were consistently strong for screening) while highlighting that genotype discrimination is the most challenging setting and may require representations optimized for subtle repolarization morphology.

**3.5 Lead Masking Analysis and Clinical Interpretability**

We performed systematic lead-group masking on the standalone LASAN model to test whether learned decision-making aligns with known electrophysiologic anatomy (**Figure 2**). Masking right precordial leads (V1–V3) produced a macro-AUROC drop of 3.17%, with a larger class-specific decrease for ARVC (4.54%) than for LQTS (2.71%) or Control (2.25%). This disproportionate degradation for ARVC indicates that right precordial information is particularly important for ARVC discrimination, consistent with the localization of characteristic abnormalities in V1–V3.

In contrast, masking lateral leads (I, V5, V6) preferentially impacted LQTS classification. Lateral masking reduced LQTS AUROC by 2.60% while minimally affecting ARVC (0.20%), with a macro-AUROC drop of 2.07% and a Control drop of 3.40%. This disease-specific dependence on lateral leads is concordant with repolarization abnormalities and T-wave morphology changes that are typically most evident in lateral distributions.

An ablation masking all precordial leads (V1–V6) produced the largest overall degradation (macro-AUROC drop 6.45%), with substantial declines across classes (ARVC 7.08%, LQTS 6.65%, Control 5.60%), supporting that precordial leads contain the dominant diagnostic signal for this multi-class inherited arrhythmia task. In comparison, masking limb leads (I, II) caused relatively modest degradation (macro-AUROC drop 0.70%). Together, these masking experiments provide quantitative interpretability evidence that LASAN relies on anatomically appropriate lead-group information for ARVC and LQTS discrimination rather than nonspecific or spurious correlates (**Figure 2**).

## 4. DISCUSSION

### 4.1 Principal Findings

This multi-center study demonstrates that lead-aware deep learning achieves clinically meaningful, state-of-the-art performance for inherited arrhythmia classification from routine ECGs. Three key findings support future clinical deployment. First, foundation models required task-specific adaptation: end-to-end fine-tuning consistently outperformed frozen feature extraction, indicating that inherited arrhythmia phenotypes benefit from refinement of pretrained representations rather than relying on general cardiac features alone. Second, LASAN's lead-aware architecture provided excellent performance across multiple integration strategies (whether trained standalone, combined with foundation encoders, or used in hybrid fusion) establishing a clinically based method that transfers effectively across different model backbones. Third, lead masking techniques validated physiologic plausibility: the model's preferential reliance on right precordial leads for ARVC and lateral leads for LQTS aligned with established electrophysiology, providing evidence that predictions are aligned with clinical signals.

### 4.2 Comparison to Previous Studies

As discussed previously, prior work on inherited arrhythmia classification has largely focused on single-disease detection. While cohorts across studies vary, our model demonstrated exceptional performance within the given cohort, warranting testing in additional cohorts across all tasks. For comparison, in LQTS, Jiang et al. achieved external validation AUROC of 0.93 for LQTS detection using task-specific deep learning, demonstrating strong discrimination in multi-site cohorts. For ARVC, the best performance achieved to date demonstrated an AUROC of 0.94 in hold-out testing.[11] Foundation models have been applied to a limited extent in inherited arrhythmias, as this is generally not a focus of classification benchmarks for these models. Notably, Deep-ECG-SSL, utilized in this analysis, has achieved the highest available AUROC 0.931 for LQTS subtype classification (LQT1 vs. LQT2).[23] Our models outperform the above studies, achieving higher AUROC for at least one architecture and foundation model combination model in all tasks, while achieving a mean AUROC across hybrid architectures of 0.985, 0.945 and 0.854 for ARVC detection, LQTS detection, and LQTS subtype differentiation respectively.

In addition to improving on binary performance, the present study advances the field through unified multi-class discrimination (ARVC vs LQTS vs control) within a single model family, a capability relevant for clinical differential diagnosis. Our hybrid LASAN integration with HuBERT-ECG achieved near-ceiling performance (AUROC 0.999 for ARVC vs control; 0.994 for LQTS vs control; 0.990 for multi-class), significantly exceeding benchmark performance while adding interpretable lead-dependence patterns. This positions lead-aware architectures as a viable path forward for deployable screening tools.

**4.3 Clinical Deployment Considerations**

The most immediate application of our work is automated triage in settings where expert interpretation may be delayed or unavailable. High-discrimination models can support family cascade screening (early identification triggers confirmatory testing and preventive therapy), pre-participation athletic screening (where missed diagnoses carry high consequence), and emergency/primary care environments (where inherited syndromes may be under-diagnosed). The interpretability results further support deployment, as predictions supported by lead-group importance patterns (V1-V3 emphasis for ARVC; lateral lead emphasis for LQTS) enable greater clinician trust, who, when using these models, can prioritize cases for specialist review more rapidly.

Overall, the 3-way classification models provide a significant step towards a unified ECG screening model for all inherited arrhythmias.In addition, in compute limited settings, smaller architectures (i.e., Standalone LASAN) can be utilized to enable rapid diagnosis without foundation model implementation. Implementation of these models in practice could follow a staged validation pathway in accordance with current AI integration guidelines.[41, 42]

**4.4 Foundation Model Integration Strategy**

While foundation models provide strong initialization through large-scale pretraining, our results indicate downstream adaptation remains essential for rare, morphologically subtle phenotypes. This finding has practical implications: frozen encoders may prove insufficient for inherited arrhythmia screening, and evaluation protocols should explicitly compare adaptation strategies rather than assuming pretrained representations alone will suffice. Clinically, this suggests that model pipelines for specialized phenotypes will benefit from using targeted fine tuning methods to capture syndrome-specific morphology, while also utilizing the advantages of pre-trained models. The hybrid LASAN approach strongly exemplifies this architecture-representation design, as pretrained encoders are paired with lead-aware processing structures to yield robust performance across backbones, suggesting a generalizable design pattern for rare electrophysiology phenotypes.

### 4.5 Limitations

Several limitations merit consideration. First, this retrospective study drew from specialized referral centers, which may not reflect unselected screening populations; prospective evaluation in broader contexts remains essential before deployment. Second, although multi-center, all sites operate within one national healthcare system and geographic region; international validation across diverse acquisition environments is required. Third, in real-world inherited arrhythmia cohorts, patients enter care through different pathways and undergo variable follow-up (e.g., family screening vs evaluation for symptoms, with differing numbers and timing of ECGs). Future work should evaluate model performance across these referral contexts and across clinically relevant comorbidity groups.

### 4.6 Future Directions

Next steps should prioritize workflow-embedded prospective evaluation, including calibration assessment and operational threshold tuning aligned with clinical objectives (e.g., high sensitivity for screening vs high specificity for diagnostic confirmation). External validation in international cohorts would address generalizability and support regulatory pathways. Expansion to other inherited syndromes (Brugada syndrome, catecholaminergic polymorphic ventricular tachycardia) would provide a more comprehensive inherited arrhythmia screening suite and ideally would include risk stratification in alignment with clinical decision pathways.

## 5. CONCLUSIONS

In a 13-center Canadian cohort, lead-aware spatial attention networks achieved high-performing and robust inherited arrhythmia classification from routine ECGs, with excellent performance across clinically relevant multi-class and binary classification tasks. Systematic evaluation of transfer learning strategies showed that downstream adaptation of foundation models is important for inherited electrophysiology phenotypes, and that combining pretrained representations with lead-aware processing can yield particularly strong discrimination. Lead masking analyses demonstrated physiologically consistent lead-group dependence for ARVC and LQTS, supporting clinical plausibility and interpretability. Together, these findings establish a scalable and clinically grounded framework for automated inherited arrhythmia screening and triage, motivating prospective validation and external testing as the next steps toward deployment.

# 6. ACKNOWLEDGMENTS

**FUNDING**

The National Hearts in Rhythm Organization is funded by grant RN380020-406814 from the Canadian Institutes of Health Research.

**DISCLOSURES / CONFLICTS OF INTEREST**

The authors have no conflicts of interest to disclose.

**TABLES**
**Table 1. Patient ECG counts by condition group.**

| Group | N (Patients) | N (ECGs) | Mean ECGs/Patient | Min ECGs/Patient | Max ECGs/Patient |
|---|---|---|---|---|---|
| ARVC | 121 | 379 | 3.13 | 1 | 30 |
| LQTS | 268 | 465 | 1.74 | 1 | 36 |
| Control | 256 | 500 | 1.95 | 1 | 9 |

**Table 2. Foundation model performance by training strategy.**

| Training Strategy | Foundation Model | AUROC | AUROC Standard Deviation | AUROC LQTS | AUROC ARVC | AUROC Control |
|---|---|---|---|---|---|---|
| Linear Probing | ECG-Founder | 0.852 | 0.056 | 0.868 | 0.877 | 0.810 |
| | Deep-ECG-SSL | 0.822 | 0.051 | 0.822 | 0.847 | 0.796 |
| | ECG-FM | 0.819 | 0.039 | 0.853 | 0.807 | 0.799 |
| | HuBERT-ECG | 0.807 | 0.035 | 0.835 | 0.779 | 0.806 |
| Linear Probing + Fine Tuning | ECG-Founder | 0.898 | 0.025 | 0.916 | 0.930 | 0.847 |
| | Deep-ECG-SSL | 0.909 | 0.014 | 0.925 | 0.930 | 0.871 |
| | ECG-FM | 0.901 | 0.020 | 0.909 | 0.925 | 0.870 |
| | HuBERT-ECG | 0.883 | 0.019 | 0.892 | 0.899 | 0.858 |
| Fine Tuning | ECG-Founder | 0.910 | 0.018 | 0.918 | 0.942 | 0.869 |
| | Deep-ECG-SSL | 0.908 | 0.013 | 0.927 | 0.922 | 0.877 |
| | ECG-FM | 0.909 | 0.025 | 0.916 | 0.929 | 0.882 |
| | HuBERT-ECG | 0.889 | 0.018 | 0.898 | 0.907 | 0.862 |

**Table 3. LASAN Model Architectures Utilizing Varied Foundational Model Components**

| Model Architecture | Foundation Model | AUROC | AUROC Standard Deviation | AUROC LQTS | AUROC ARVC | AUROC Control |
|---|---|---|---|---|---|---|
| Standalone LASAN Architecture | None | 0.911 | 0.037 | 0.932 | 0.940 | 0.862 |
| Foundation Model with LASAN Head | ECG-Founder | 0.827 | 0.097 | 0.852 | 0.907 | 0.722 |
| | Deep-ECG-SSL | 0.909 | 0.014 | 0.925 | 0.930 | 0.871 |
| | ECG-FM | 0.973 | 0.004 | 0.983 | 0.971 | 0.964 |
| | HuBERT-ECG | 0.990 | 0.003 | 0.989 | 0.994 | 0.987 |
| Hybrid LASAN Architecture | ECG-Founder | 0.950 | 0.011 | 0.954 | 0.954 | 0.941 |
| | Deep-ECG-SSL | 0.932 | 0.016 | 0.930 | 0.945 | 0.920 |
| | ECG-FM | 0.975 | 0.003 | 0.983 | 0.975 | 0.967 |
| | HuBERT-ECG | 0.990 | 0.002 | 0.989 | 0.994 | 0.99 |

**Table 4. Architecture Performance on Binary Classification Tasks**

| Task | Model Architecture | Foundation Model | AUROC | Sensitivity | Specificity | Accuracy |
|---|---|---|---|---|---|---|
| ARVC vs. Control | Standalone LASAN Architecture | None | 0.974 | 1 | 0.864 | 0.941 |
| | Hybrid LASAN Architecture | ECG-Founder | 0.978 | 0.955 | 0.924 | 0.941 |
| | | Deep-ECG-SSL | 0.984 | 0.870 | 0.949 | 0.904 |
| | | ECG-FM | 0.991 | 0.864 | 0.966 | 0.908 |
| | | HuBERT-ECG | 0.999 | 0.961 | 0.983 | 0.971 |
| LQTS vs. Control | Standalone LASAN Architecture | None | 0.901 | 0.701 | 0.939 | 0.811 |
| | Hybrid LASAN Architecture | ECG-Founder | 0.939 | 0.870 | 0.848 | 0.860 |
| | | Deep-ECG-SSL | 0.930 | 0.818 | 0.720 | 0.773 |
| | | ECG-FM | 0.962 | 0.812 | 0.955 | 0.878 |
| | | HuBERT-ECG | 0.994 | 0.948 | 0.970 | 0.958 |
| LQTS Type 1 vs. LQTS Type 2 | Standalone LASAN Architecture | None | 0.920 | 0.955 | 0.650 | 0.758 |
| | Hybrid LASAN Architecture | ECG-Founder | 0.764 | 0.477 | 0.812 | 0.694 |
| | | Deep-ECG-SSL | 0.948 | 0.977 | 0.650 | 0.766 |
| | | ECG-FM | 0.738 | 0.523 | 0.700 | 0.637 |
| | | HuBERT-ECG | 0.901 | 0.841 | 0.775 | 0.798 |

# FIGURES
## Figure 1. Illustration of LASAN Architectures

**Figure 1a. LASAN Architecture.** Illustration of the components of the LASAN method architecture, applies lead-aware attention to lead-specific and anatomically based position encodings to produce predictions. Lead Importance weights can be extracted from the attention mechanism to assess clinical interpretability.

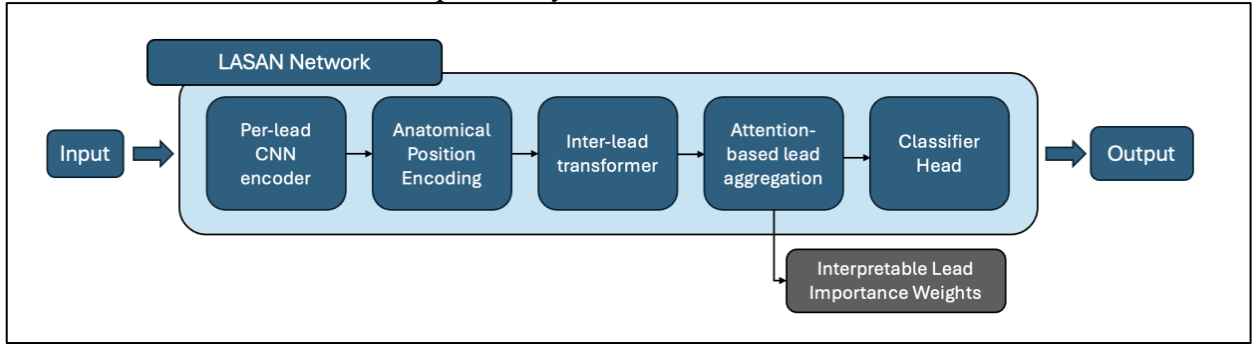

**Figure 1b. Standalone LASAN Architecture.** Illustration of Standalone LASAN architecture, which uses only the LASAN model to produce prediction and does not incorporate foundation model embeddings.

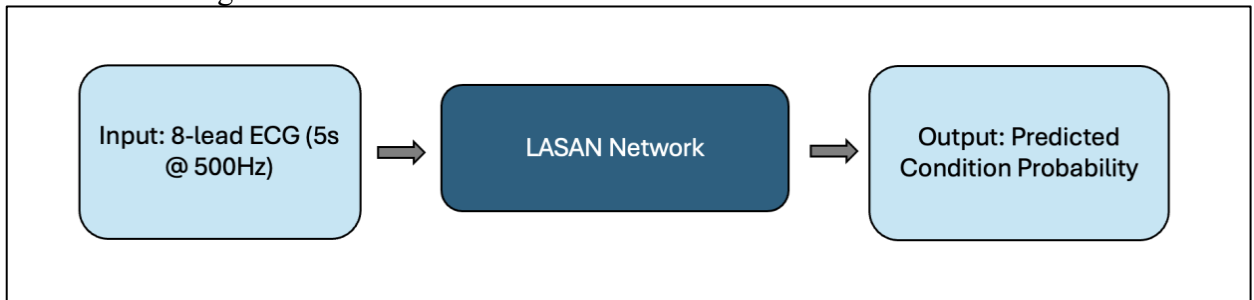

**Figure 1c. Foundation Model + LASAN Head Architecture.** Illustration of Foundation Model + LASAN Head, which fuses the embeddings of a LASAN model and foundation model to produce predictions.

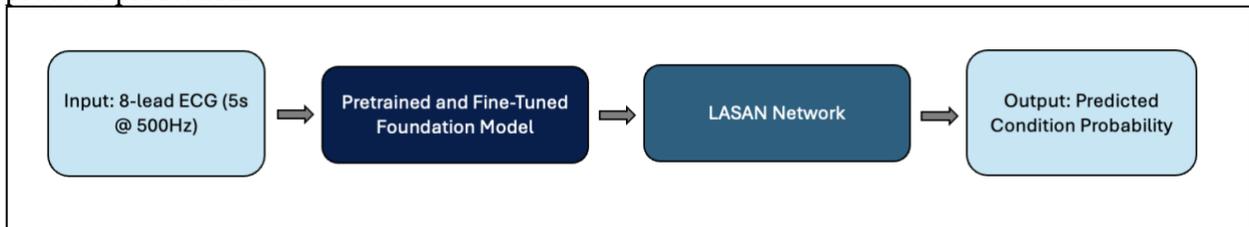

**Figure 1d. Hybrid Model Architecture.** Illustration of LASAN Hybrid Model Architecture, which fuses the embeddings of a LASAN model and foundation model to produce predictions.

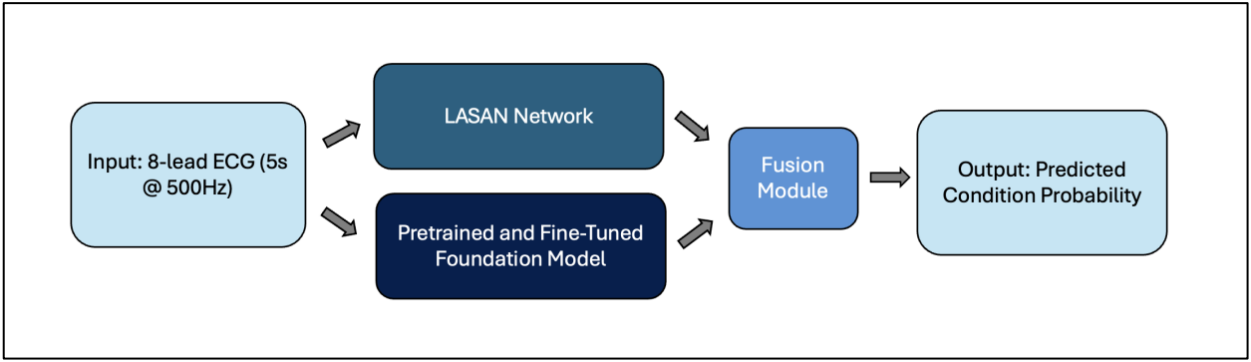

**Figure 2. Disease-specific lead importance revealed by group masking analysis. AUROC drop (%) when specific lead groups are masked during inference**. Right precordial leads (V1-V3) demonstrate greater importance for ARVC detection compared to LQTS, while lateral leads (I, V5, V6) are more important for LQTS compared to ARVC. Data sourced from LASAN standalone model (AUROC = 0.911) for multi-classification (ARVC vs. LQTS vs. Control).

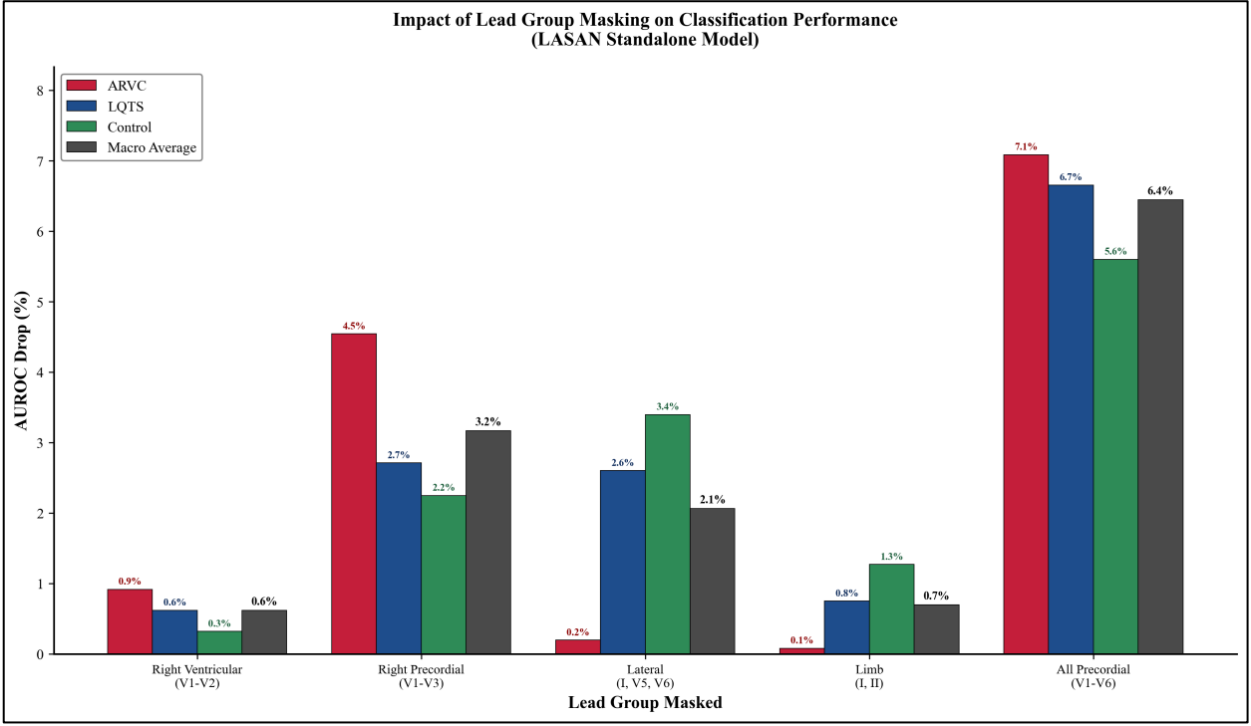